\documentclass{article}

\usepackage{microtype}
\usepackage{graphicx}
\usepackage{subcaption}
\usepackage{booktabs} 
\usepackage{listings}

\usepackage{hyperref}

\usepackage[preprint]{icml2026}

\usepackage{amsmath}
\usepackage{amssymb}
\usepackage{mathtools}
\usepackage{amsthm}
\usepackage{adjustbox}
\usepackage{multirow}
\usepackage{colortbl}
\usepackage{nicematrix}

\def\eg{\emph{e.g.}}

\def\fig{Fig. }
\def\tab{Tab. }

\definecolor{icmlblue}{HTML}{DCEBFF} 
\definecolor{mygray}{gray}{0.6}
\newcommand{\name}[1]{ToolTok}

\usepackage{enumerate}
\usepackage{enumitem}

\usepackage[capitalize,noabbrev]{cleveref}

\theoremstyle{plain}

\theoremstyle{definition}

\theoremstyle{remark}

\usepackage[textsize=tiny]{todonotes}

\icmltitlerunning{ToolTok: Tool Tokenization for Efficient and Generalizable GUI Agents}

\begin{document}

\twocolumn[
  \icmltitle{\name{}: Tool Tokenization for Efficient and Generalizable GUI Agents}
  
  \icmlsetsymbol{equal}{*}

  \begin{icmlauthorlist}
    \icmlauthor{Xiaoce Wang}{thu}
    \icmlauthor{Guibin Zhang}{nus}
    \icmlauthor{Junzhe Li}{pku}
    \icmlauthor{Jinzhe Tu}{thu}
    \icmlauthor{Chun Li}{bit}
    \icmlauthor{Ming Li}{gml}
  \end{icmlauthorlist}

  \icmlaffiliation{thu}{Department of Computer Science, Tsinghua University, Beijing, China}
  \icmlaffiliation{gml}{Guangming Laboratory, Shenzhen, China}
  \icmlaffiliation{nus}{National University of Singapore, Singapore}
  \icmlaffiliation{pku}{Peking University, Beijing, China}
  \icmlaffiliation{bit}{MSU-BIT-SMBU Joint Research Center of Applied Mathematics, Shenzhen MSU–BIT University}

  \icmlcorrespondingauthor{Ming Li}{ming.li@u.nus.edu}
  \icmlkeywords{GUI agent, cross-modal learning}

  \vskip 0.3in
]

\printAffiliationsAndNotice{}

\begin{abstract}
Existing GUI agent models relying on coordinate-based \textit{one-step} visual grounding struggle with generalizing to varying input resolutions and aspect ratios. Alternatives introduce coordinate-free strategies yet suffer from learning under severe data scarcity. To address the limitations, we propose \name{}, a novel paradigm of \textit{multi-step} pathfinding for GUI agents, where operations are modeled as a sequence of progressive tool usage. Specifically, we devise tools aligned with human interaction habits and represent each tool using learnable token embeddings. To enable efficient embedding learning under limited supervision, \name{} introduces a semantic anchoring mechanism that grounds each tool with semantically related concepts as natural inductive bias. To further enable a pre-trained large language model to progressively acquire tool semantics, we construct an easy-to-hard curriculum consisting of three tasks: token definition question-answering, pure text-guided tool selection, and simplified visual pathfinding.  Extensive experiments on multiple benchmarks show that \name{} achieves superior performance among models of comparable scale (4B) and remains competitive with a substantially larger model (235B). Notably, these results are obtained using \textbf{less than 1\%} of the training data required by other post-training approaches. In addition, \name{} demonstrates strong generalization across unseen scenarios. Our training \& inference code is open-source at \hyperlink{https://github.com/ZephinueCode/ToolTok}{code}.
\end{abstract}
\vspace{-0.3em}
\section{Introduction}
\vspace{-0.3em}
\begin{figure}[ht]
  \begin{center}
    \centerline{\includegraphics[width=\linewidth]{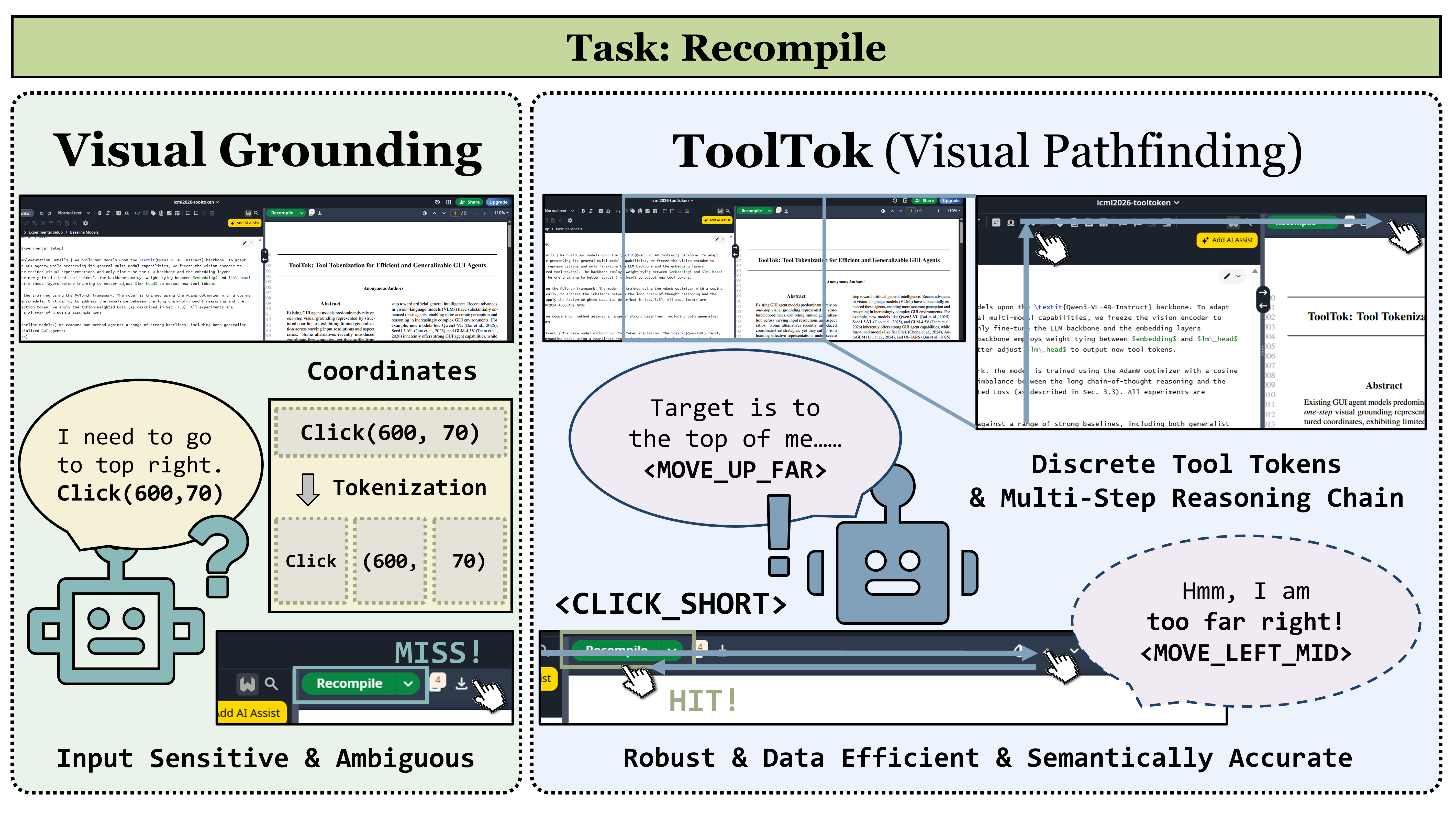}}
    \caption{
      Comparison of GUI agent interaction paradigms. Left: One-step visual grounding with direct coordinate prediction. Right: \textbf{ToolTok}, which performs multi-step visual pathfinding via discrete tool tokens, leading to improved robustness, data efficiency, and semantic interpretability.
    }
    \label{fig:teaser}
  \end{center}
  \vskip -0.3in
\end{figure}

General-purpose agents capable of interacting with graphical user interfaces (GUIs) are widely regarded as a critical step toward artificial general intelligence. Recent advances in vision–language models (VLMs) have substantially enhanced these agents, enabling more accurate perception and reasoning in increasingly complex GUI environments. For example, new models like Qwen3-VL \cite{bai2025qwen3vltechnicalreport}, Seed1.5-VL \cite{guo2025seed15vltechnicalreport}, and GLM-4.5V \cite{vteam2026glm45vglm41vthinkingversatilemultimodal} inherently offer strong GUI agent capabilities, while fine-tuned models like SeeClick \cite{cheng2024seeclick}, AutoGLM \cite{liu2024autoglm}, and UI-TARS \cite{qin2025ui} provide more specialized options.

The dominant paradigm for current GUI agents is \textit{one-step} visual grounding \cite{nguyen2025gui}, where agents execute actions by directly generating structured coordinates or bounding boxes, \eg, \textit{[click, [0.5, 0.5]]}. This formulation reduces GUI interaction to explicit numeric prediction, rather than a semantic decision-making process akin to human behavior.
To support coordinate prediction, such models assume a fixed global coordinate system during training, which in turn requires all input screenshots to be normalized to a predefined resolution and aspect ratio at inference time. This design introduces two key issues.
First, normalization inevitably alters the original visual content of GUI screenshots: rescaling and downsampling may distort fine-grained interface elements, such as button shapes or layout details, thereby degrading perceptual cues.
Second, it tightly couples learned action representations with the input format, resulting in fragile behavior when test-time resolutions or aspect ratios deviate from those seen during training.

\begin{figure}[ht]
  \vspace{-10pt}
  \begin{center}
    \centerline{\includegraphics[width=\linewidth]{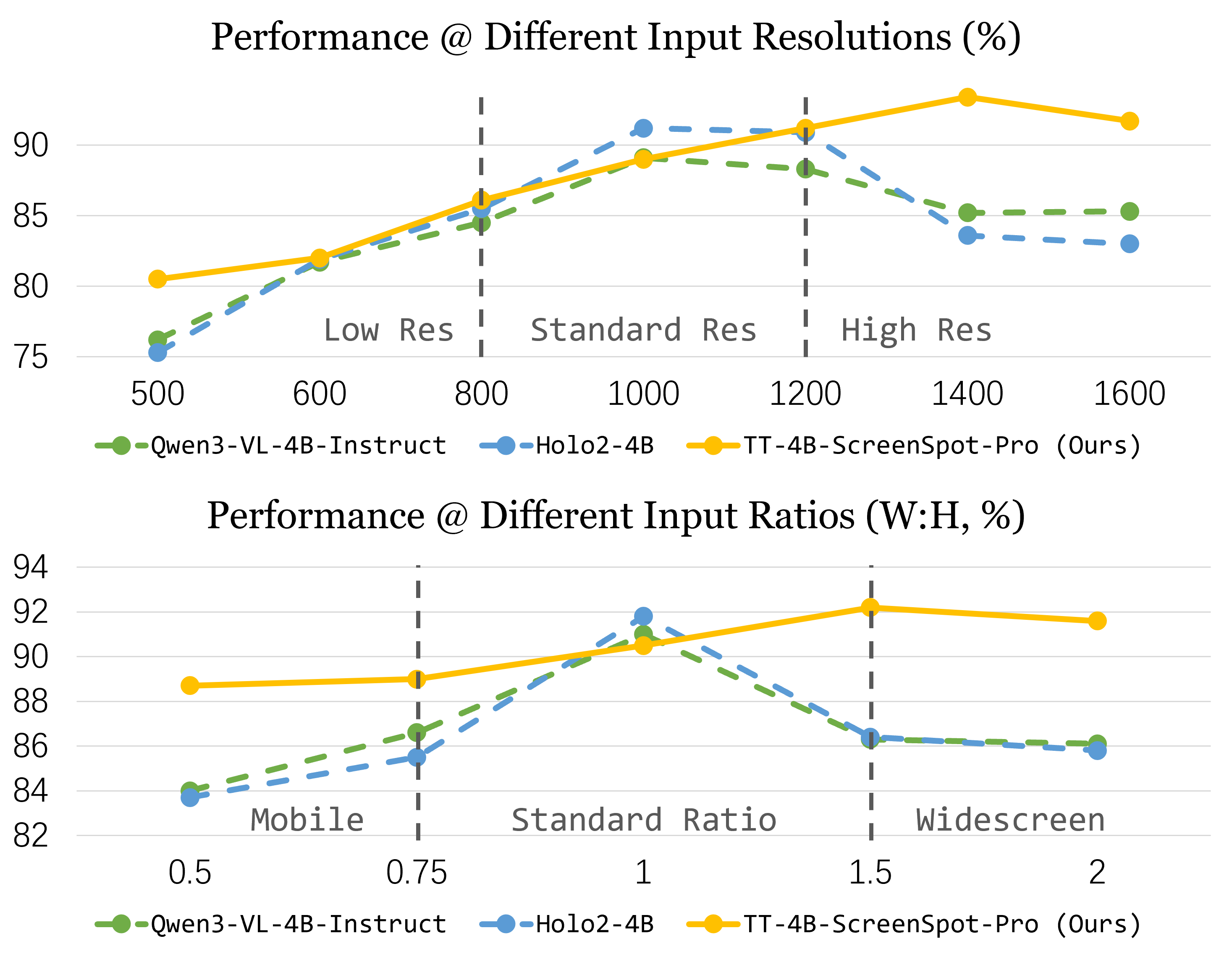}}
    \caption{
    Performance comparison under varying resolutions and aspect ratios. While baselines degrade significantly in extreme cases, our model demonstrates superior robustness using \textbf{less than 1\%} of the training data.
    }
    \label{fig:robustness}
  \end{center}
  \vskip -0.25in
\end{figure}

As a consequence, grounding-based GUI agents exhibit pronounced sensitivity to screenshot settings, such as variations in input resolution and aspect ratio.
We empirically validate this phenomenon in \fig \ref{fig:robustness} by evaluating two existing agents on ScreenSpot \cite{cheng2024seeclick} under varying input resolutions and aspect ratios. Performance consistently peaks when the test configuration closely matches the training settings, and degrades substantially as these conditions deviate.
These results confirm that coordinate-based visual grounding struggles to generalize across diverse GUI layouts.

\vspace{-0.3em}

Alternative approaches, such as TAG \cite{wu2025gui} and GUI-Actor \cite{xu2025attention}, attempt to mitigate these problems by employing coordinate free methods. These models primarily utilize attention maps or dedicated tokens that attend to certain visual patches. Therefore, they align agentic tasks more efficiently with the VLM backbone's pre-trained priors, thus improving robustness. However, these methods remain fundamentally limited, as they primarily predict screen locations rather than explicitly output specific GUI actions. Not only does this confine these models to simple ``point and click'' problems (e.g., ``Find the search bar on the screen'') instead of a larger variety of possible GUI tasks, but it also means that a large amount of natural language knowledge of pre-trained VLMs is left not effectively leveraged. As a result, these methods require voluminous training data only for limited utility. Refer to Appendix \ref{app:related_works} for related work details.

In this work, we rethink GUI interaction in a completely different manner and focus on a question: \textit{Can we define learnable tools that imitate real human behaviors in GUI interactions}? The primary challenge lies in representation alignment under data scarcity. Introducing new, randomly initialized tool tokens and optimizing them from scratch within a pre-trained VLM leads to a severe ``cold-start'' problem \cite{kim2024initializing}, especially when training data are limited. Prior studies \cite{peng2023kosmos, liu2023visual} have shown that aligning newly introduced visual or semantic concepts typically requires large, carefully curated datasets. This challenge is particularly acute for GUI agents, where collecting high-quality interaction trajectories at scale is inherently difficult. Existing public benchmarks are relatively small, containing fewer than 2,000 samples. For example, ScreenSpot includes approximately 1,300 samples, and ScreenSpot-Pro \cite{li2025screenspot} contains around 1,600 samples, rendering naive supervised fine-tuning insufficient for effective tool alignment.

To this end, we propose \textbf{\name{}}, a novel paradigm that shifts GUI agents from \textit{one-step visual grounding} to \textit{multi-step progressive pathfinding}. Instead of formulating GUI interaction as continuous coordinate regression, we recast the problem as iteratively determining \emph{which discrete tool enables progress toward the goal}, thereby framing GUI control as a classification problem over a structured tool space. By discretizing interactions into callable tool tokens, \name{} more closely mirrors human behavior in navigating graphical interfaces.

To address representation alignment under data scarcity, we introduce a semantic anchoring mechanism that grounds each tool in semantically related concepts rather than treating it as an uninformative symbol. Concretely, each tool is associated with a small set of function descriptive words. For example, the tool $\texttt{<MOVE\_UP\_FAR>}$ is linked to the concepts $\{move, up, far\}$. By embedding tools within the model’s existing semantic space, our mechanism provides a natural inductive bias that allows the model to interpret and reason about tools through its pre-trained knowledge. This design significantly eases the alignment of tool semantics under limited supervision, leading to faster and more stable learning in data-scarce regimes.

To further bridge linguistic knowledge with actionable tool understanding, we structure the optimization process as a curriculum learning trajectory that reflects the gradual acquisition of tool concepts. The model is initially guided to reason about tools at a purely semantic level, through language-centric supervision including \textit{token definition question-answering} and \textit{text-guided tool selection}, before being progressively introduced to decision-making and perception-grounded settings via \textit{simplified visual pathfinding}. By gradually strengthening the interaction among language, tools, and visual cues, this learning process enables the model to internalize tool semantics in a stable and data-efficient manner, requiring merely 5,000 synthetic samples.

We conduct extensive experiments on multiple widely used GUI benchmarks, showing that \name{} consistently outperforms existing methods even without relying on large-scale pre-training data. \name{} achieves state-of-the-art (SOTA) performance among models of comparable scale (4B) and delivers competitive results compared to substantially larger models (235B). Notably, \name{} exhibits markedly stronger generalization to variations in input resolution and aspect ratio.

Our main contributions are summarized as follows:
\vspace{-0.3em}
\begin{itemize}[leftmargin=1em, itemsep=0pt, topsep=0pt]
    \item We propose a new paradigm that shifts GUI agents from visual grounding to progressive visual pathfinding, enabling interaction through discrete learnable tools that better reflect human-like GUI navigation.
    
    \item We introduce semantic anchoring, a principled mechanism that ties newly introduced tools to learned semantic knowledge, facilitating effective and stable representation alignment under limited supervision.
    
    \item We design a curriculum learning scheme that gradually couples language, tools, and visual perception, substantially alleviating data scarcity in fine-tuning GUI agents.
\end{itemize}

\section{Method}

\begin{figure*}[ht]
  \begin{center}
    \centerline{\includegraphics[width=0.99\textwidth]{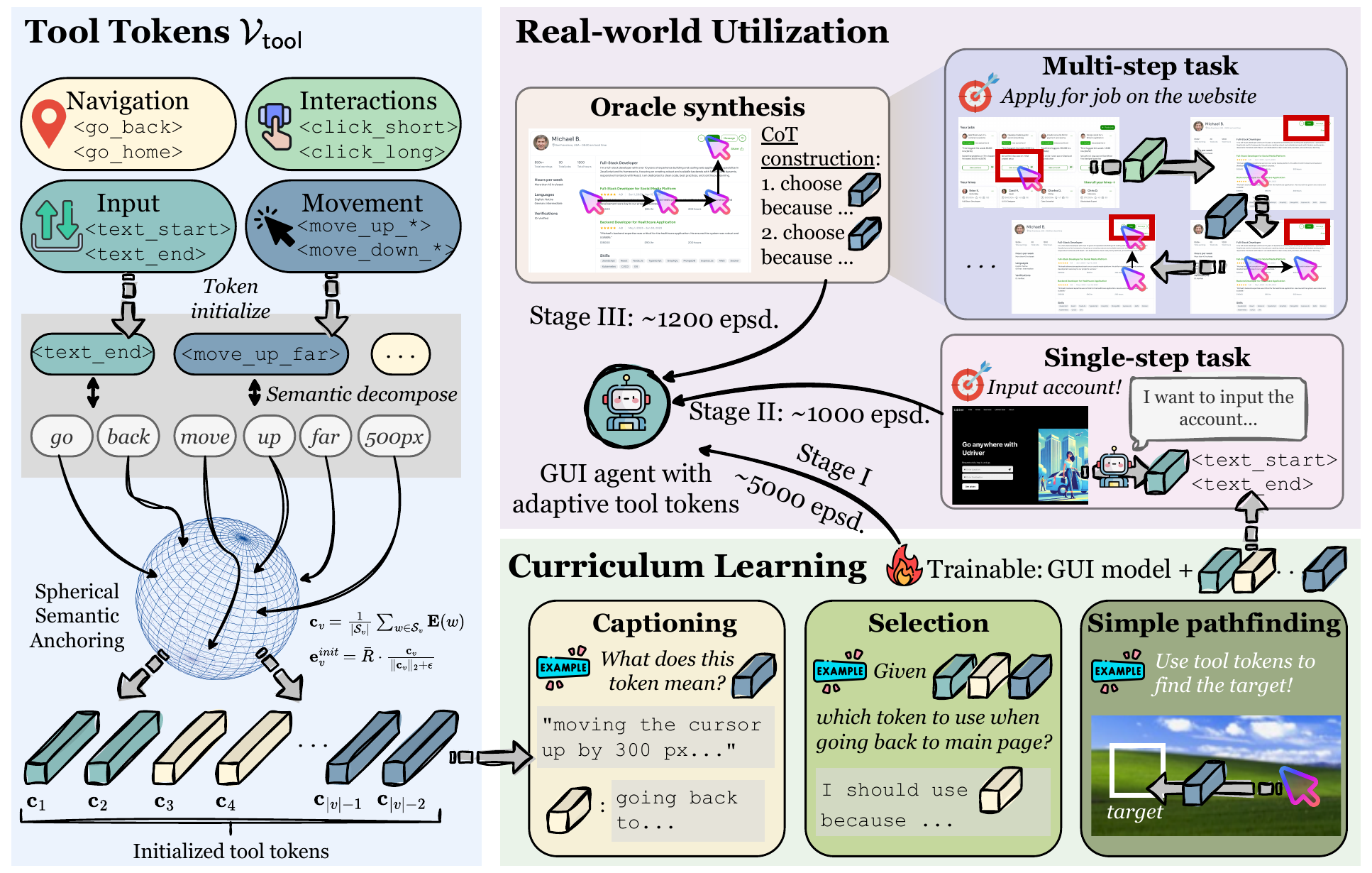}}
    \caption{
      The main pipeline of \name{}. Featuring discrete action tokens, an efficient token initialization method of spherical semantic anchoring, and a curriculum learning schedule featuring three stages.
    }
    \label{fig:main}
  \end{center}
  \vskip -0.3in
\end{figure*}

\subsection{Preliminary}
\vspace{-0.5em}
\paragraph{Problem Formulation.} 
We model the GUI navigation task as a sequential decision process. At each time step $t$, the agent receives a natural language instruction $x$ and observes the current GUI state. Drawing inspiration from human visual interaction, where one tracks the cursor visually rather than calculating coordinates, we propose a \textbf{visually augmented state}. Instead of relying on a raw screenshot $I_t$, we inject an explicit visual marker (e.g., a crosshair) at the agent's current implicit cursor position $p_t$ to form the observation $o_t = (\tilde{I}_t, x, h_{t-1})$. Here, $\tilde{I}_t$ represents the rendered screenshot with visual feedback, and $h_{t-1}$ denotes the interaction history. This design relieves the model from memorizing absolute coordinates, allowing it to focus on relative visual guidance.

\vspace{-0.5em}
\paragraph{Generative Decision Process.} 
We aim to learn a policy $\pi_\theta$ that generates the optimal action sequence. 
Given the augmented observation $o_t$, the agent acts as an autoregressive generator, producing a Chain-of-Thought (CoT) reasoning path $r_t$ followed by an executable action token $a_t \in \mathcal{V}_{tool}$. 
The joint probability is factorized as:
\begin{equation}
    \pi_\theta(r_t, a_t | o_t) = \prod_{i=1}^{|r_t|} p(r_{t,i} | r_{t,<i}, o_t) \cdot p(a_t | r_t, o_t).
\end{equation}
By maximizing this likelihood, the model learns to ground its reasoning in the visual state $\tilde{I}_t$ before committing to a specific GUI operation.

\vspace{-0.5em}
\subsection{Introducing New Tool Tokens}
\vspace{-0.3em}
\paragraph{Discretized Action Space.} 
To address the ambiguity of continuous coordinate generation, we construct a discrete action vocabulary $\mathcal{V}_{tool}$. Our vocabulary consists of semantic tokens categorized into four functional groups: 
\begin{itemize}[leftmargin=1em, itemsep=0pt, topsep=0pt]
\item \textbf{Movement:} $\texttt{<MOVE\_UP\_*>}$, $\texttt{<MOVE\_DOWN\_*>}$, $\texttt{<MOVE\_LEFT\_*>}$, $\texttt{<MOVE\_RIGHT\_*>}$. Used to move the cursor on screen. $*$ is the distance of the action.
\item \textbf{Navigation:} $\texttt{<GO\_BACK>}$, $\texttt{<GO\_HOME>}$. Used for navigating in the operating system.
\item \textbf{Interactions:} $\texttt{<CLICK\_SHORT>}$, $\texttt{<CLICK\_LONG>}$, $\texttt{<SCROLL\_UP>}$ (and $\texttt{DOWN}$, $\texttt{LEFT}$, and $\texttt{RIGHT}$). Used for interacting with elements on screen.
\item \textbf{Input:} $\texttt{<TEXT\_START>}$, $\texttt{<TEXT\_END>}$. Used for indicating the start and the end of text input.
\end{itemize}

The exact definition for each token can be found in Appendix \ref{app:token_definition}. This formulation ensures that given relevant data, the system can perform most common GUI tasks, and that the semantics of each token is well-defined, which allows us to easily align the action space with the pre-trained token space of the VLM backbone. 

\vspace{-0.3em}
\paragraph{Hierarchical Action Decoding.}
Cursor movement is the key to nearly all GUI tasks. To balance the trade-off between navigation efficiency and precision, we structure cursor movement tokens in $\mathcal{V}_{tool}$ as a hierarchical system. These tokens are decomposed into direction and scale: $a_{mv} = \langle d, s \rangle$, where $d \in \{\uparrow, \downarrow, \leftarrow, \rightarrow\}$ and $s \in \{\texttt{FAR}, \texttt{MID}, \texttt{CLO}\}$. 
During inference, the execution engine translates a generated token into a coordinate delta $\Delta p$:
\begin{equation}
    \Delta p = \delta(d) \cdot \mu(s), \quad \text{where } \mu(s) \in \{500, 150, 30\}.
\end{equation}
$\texttt{FAR}$ translates to $500$ pixels of movement, whereas $\texttt{MID}$ and $\texttt{CLO}$ translates to $150$ and $30$ pixels respectively. This hierarchical design encourages a \textit{coarse-to-fine} search strategy: the agent first traverses large distances using \texttt{FAR} tokens and then refines its position with \texttt{CLO} tokens, mimicking human cursor control behavior. The comprehensive list of all tool tokens is provided in Appendix \ref{app:token_definition}.

\vspace{-0.3em}
\paragraph{Spherical Semantic Anchoring.}
A critical challenge in extending the vocabulary of a pre-trained VLM is the \textit{semantic mismatch} between randomly initialized new tokens and the well-structured pre-trained embedding space. 
To address this, we introduce \textbf{Spherical Semantic Initialization (SSI)}. 
Let $\mathbf{E}(\cdot)$ denote the embedding function of the pre-trained model. For each tool token $v \in \mathcal{V}_{tool}$, we first curate a set of natural language anchors $\mathcal{S}_v$ that describe its function (e.g., for \texttt{<MOVE\_UP\_FAR>}, $\mathcal{S}_v = \{\text{``move''}, \text{``up''}, \text{``north''}, \text{``far''}\}$).
Previous studies suggest that pre-trained embeddings largely fall into a narrow cone or hypersphere in high-dimensional space \cite{gao2019representation}. Therefore, simple average of the embedding of any tokens will likely land outside the semantically meaningful space due to its \textbf{norm} being expectedly smaller. To counter this, we compute the arithmetic centroid $\mathbf{c}_v$ of anchor embeddings and project it onto the hypersphere surface defined by the average norm $\bar{R}$ of the original vocabulary:

\begin{equation}
    \mathbf{c}_{v} = \frac{1}{|\mathcal{S}_{v}|} \sum_{w \in \mathcal{S}_{v}} \mathbf{E}(w), \quad
    \mathbf{e}_{v}^{init} = \bar{R} \cdot \frac{\mathbf{c}_{v}}{\| \mathbf{c}_{v} \|_2 + \epsilon},
\end{equation}

where $\epsilon$ is a small constant (e.g., $10^{-8}$) added for numerical stability. This initialization more effectively \textit{anchors} the tool tokens to their semantic neighbors in the latent space, enabling the model to better and faster align the tool tokens without damaging pre-trained language capabilities.

\vspace{-0.3em}
\subsection{Curriculum-based ToolTok Learning}
\vspace{-0.3em}
\label{sec: training_stategy}
Although these GUI tool tokens are well initialized, agents are not guaranteed to internalize their underlying dynamics or to exploit them fluently in GUI interactions. We therefore design a data-centric curriculum that progressively grounds the agent’s understanding, advancing from \textit{token semantic alignment} to \textit{simplified pathfinding} and finally \textit{complex real-world navigation}. This requires us to formulate effective training strategies for both synthetic and real-world data.

\subsubsection{Synthetic Data Training}

\vspace{-0.3em}
\paragraph{Data Synthesis.} Before moving to high quality, real-world datasets, we first need to train the agent to further internalize the newly initialized GUI tool tokens using easier synthetic tasks. This stage aims to ensure that the agent acquires a precise semantic grounding of the newly introduced tool tokens and understands their functional roles in GUI manipulation. Accordingly, we synthesize tasks that explicitly comprehend and utilize these tokens. Concretely, we construct a noise-free synthetic dataset $\mathcal{D}_{\mathrm{syn}}$ comprising three distinct tasks to rapidly align the tool tokens with their intended semantics: \textbf{Token definition QA}, \textbf{text-guided tool selection}, and \textbf{simplified visual pathfinding}. We construct a total of $5,000$ synthetic data points. Refer to Appendix \ref{app:data_synth} for detailed definition and size of each type of synthesized data.

\vspace{-0.3em}
\paragraph{Stage I Training.} As the first step of the training process, we train the model using standard next token loss \cite{bengio2003neural} on $\mathcal{D}_{syn}$. This stage acts as an advanced ``warm-up'', establishing the basic syntax of $\mathcal{V}_{tool}$ and spatial logic. It ensures that when the agent later sees real images, it already ``knows'' how to move, reducing the complexity of learning more complex GUI tasks.

\subsubsection{Real-World Dataset Training}

\vspace{-0.3em}
\paragraph{Trajectory Synthesis.}
After stage I training, we consider training the agent with real-world data. However, standard real-world GUI datasets typically provide only the static target bounding box $B_{gt}$ without interaction traces. This single-step nature does not match our multi-step approach. To enable multi-step trajectory-level supervision, we propose an \textbf{oracle trajectory synthesis} mechanism. Given a starting cursor position $p_0$ (initialized randomly) and a target $B_{gt}$, we construct a greedy shortest path $\tau^* = (a_1, a_2, \dots, a_T)$ using our discrete action space. At each step $t$, the oracle selects the optimal action $a_t \in \mathcal{V}_{tool}$ that minimizes the Euclidean distance to the target center, subject to the constraint that the action does not overshoot $B_{gt}$ unless necessary. This transforms static \textit{(Image, BBox)} pairs into multi-step supervision sequences:
\begin{equation}
    \mathcal{D}_{traj} = \{ (\tilde{I}_t, x, r_t^*, a_t^*) \}_{t=1}^{T},
\end{equation}

\paragraph{Chain-of-Thought Construction.} Chain-of-thought is proven to significantly improve data efficiency, especially for smaller models, when faced with complex tasks \cite{gunasekar2023textbooks, hsieh2023distilling, mukherjee2023orca}. Because large amounts of high-quality GUI data is difficult to obtain, when constructing trajectories, we introduce a procedurally-generated reasoning chain $r_t^*$ for each step. Detailed implementation of CoT is in Appendix \ref{app:cot}.

\begin{table*}[ht]
    \caption{Comprehensive robustness evaluation across all benchmarks. We report the standard accuracy (\textbf{Acc}) on original datasets, alongside performance under varying resolutions (\textbf{Lo}: $600$px, \textbf{Mi}: $1000$px, \textbf{Hi}: $1400$px) and aspect ratios (\textbf{Mo}: Mobile, $0.5:1$, \textbf{St}: Standard, $1:1$, \textbf{Wi}: Widescreen, $2:1$) to assess robustness. Robustness test samples undergo resizing and/or selection, therefore the average of robustness results is \textbf{not} the same as \textbf{Acc}. Refer to Appendix.~\ref{app:experiment_details} for the exact test settings. Overall, our models demonstrate superior performance and stability.}
    \label{tab:main_results}
    \begin{center}
        \begin{scriptsize}
            \begin{sc}
                \setlength{\tabcolsep}{0pt} 
                
                \begin{NiceTabular*}{0.99\textwidth}{l@{\extracolsep{\fill}}c | cccc ccc | cccc ccc}
                    \toprule
                    \multicolumn{2}{c}{} & \multicolumn{7}{c}{\textbf{ScreenSpot (In-Domain)}} & \multicolumn{7}{c}{\textbf{ScreenSpot-Pro (In-Domain)}} \\
                    \cmidrule(lr){3-9} \cmidrule(lr){10-16}
                    
                    \textbf{Model} & \textbf{Size} 
                    & \textbf{Acc$\uparrow$} & \color{gray}Lo$\uparrow$ & \color{gray}Mi$\uparrow$ & \color{gray}Hi$\uparrow$ & \color{gray}Mo$\uparrow$ & \color{gray}St$\uparrow$ & \color{gray}Wi$\uparrow$ 
                    & \textbf{Acc$\uparrow$} & \color{gray}Lo$\uparrow$ & \color{gray}Mi$\uparrow$ & \color{gray}Hi$\uparrow$ & \color{gray}Mo$\uparrow$ & \color{gray}St$\uparrow$ & \color{gray}Wi$\uparrow$ \\
                    \midrule
                    
                    Qwen3-VL-4B & 4B    
                    & 88.0 & 81.7 & 87.3 & 83.1 & 84.1 & 91.0 & 86.1 
                    & 46.6 & 29.0 & 34.9 & 46.2 & 40.4 & 47.8 & 45.0 \\
                    
                    Qwen3-VL-8B & 8B    
                    & 90.3 & \underline{82.5} & \textbf{90.4} & \underline{86.2} & 86.4 & \textbf{92.1} & \underline{89.2}
                    & 51.4 & 31.3 & 39.7 & 51.6 & 44.6 & 52.3 & 50.9 \\
                    
                    \color{gray}Qwen3-VL-235B-PF & \color{gray}235B
                    & \color{gray}31.1 & \color{gray}- & \color{gray}- & \color{gray}- & \color{gray}- & \color{gray}- & \color{gray}-
                    & \color{gray}7.5  & \color{gray}- & \color{gray}- & \color{gray}- & \color{gray}- & \color{gray}- & \color{gray}- \\

                    \color{gray}Qwen3-VL-235B & \color{gray}235B
                    & \color{gray}92.0 & \color{gray}- & \color{gray}- & \color{gray}- & \color{gray}- & \color{gray}- & \color{gray}-
                    & \color{gray}60.3 & \color{gray}- & \color{gray}- & \color{gray}- & \color{gray}- & \color{gray}- & \color{gray}- \\
                    
                    \midrule
                    
                    Holo2-4B & 4B 
                    & \underline{90.8} & 81.9 & 89.3 & 82.5 & 83.7 & \underline{91.8} & 85.6
                    & \underline{57.9} & \underline{41.0} & \underline{52.3} & \underline{61.9} & \underline{54.1} & \underline{62.1} & \underline{59.7} \\
                    
                    GUI-Actor-7B & 7B 
                    & 87.9 & 80.2 & 88.1 & 79.9 & 82.9 & 90.4 & 84.1
                    & 44.6 & 37.4 & 42.3 & 45.1 & 38.6 & 44.7 & 42.5 \\
                    
                    \midrule
                    \rowcolor{icmlblue}
                    \textbf{TT-4B-ScreenSpot} & \textbf{4B}
                    & 87.6 & \textbf{83.1} & \underline{89.6} & 84.2 & \underline{87.4} & 90.7 & 85.8
                    & 50.2 & 34.7 & 46.1 & 52.5 & 48.0 & 51.5 & 50.8 \\
                    
                    \rowcolor{icmlblue}
                    \textbf{TT-4B-ScreenSpot-Pro} & \textbf{4B}
                    & \textbf{91.8} & 82.0 & 88.8 & \textbf{93.5} & \textbf{89.0} & 90.5 & \textbf{92.2}
                    & \textbf{61.1} & \textbf{49.2} & \textbf{58.9} & \textbf{63.4} & \textbf{56.5} & \textbf{62.7} & \textbf{61.6} \\
                    
                    \bottomrule
                    \addlinespace[5pt]
                    
                    \multicolumn{2}{c}{} & \multicolumn{7}{c}{\textbf{Mind2Web-S (Out-of-Domain)}} & \multicolumn{7}{c}{\textbf{ScreenSpot-v2 (Out-of-Domain)}} \\
                    \cmidrule(lr){3-9} \cmidrule(lr){10-16}
                    
                    \textbf{Model} & \textbf{Size} 
                    & \textbf{Acc$\uparrow$} & \color{gray}Lo$\uparrow$ & \color{gray}Mi$\uparrow$ & \color{gray}Hi$\uparrow$ & \color{gray}Mo$\uparrow$ & \color{gray}St$\uparrow$ & \color{gray}Wi$\uparrow$ 
                    & \textbf{Acc$\uparrow$} & \color{gray}Lo$\uparrow$ & \color{gray}Mi$\uparrow$ & \color{gray}Hi$\uparrow$ & \color{gray}Mo$\uparrow$ & \color{gray}St$\uparrow$ & \color{gray}Wi$\uparrow$ \\
                    \midrule
                    
                    Qwen3-VL-4B & 4B 
                    & 26.4 & 19.8 & 26.7 & 26.9 & 20.6 & 29.1 & 23.2 
                    & 86.4 & 80.9 & 86.5 & 82.8 & 83.0 & 90.1 & 87.4 \\
                    
                    Qwen3-VL-8B & 8B 
                    & \underline{38.8} & \underline{26.3} & \underline{39.1} & \underline{38.0} & \underline{29.4} & \underline{39.1} & \underline{36.9} 
                    & \textbf{91.7} & 83.2 & \textbf{92.6} & 88.1 & \underline{87.0} & \textbf{92.3} & \textbf{90.4} \\

                    \color{gray}Qwen3-VL-235B-PF & \color{gray}235B
                    & \color{gray}11.0 & \color{gray}- & \color{gray}- & \color{gray}- & \color{gray}- & \color{gray}- & \color{gray}-
                    & \color{gray}38.0 & \color{gray}- & \color{gray}- & \color{gray}- & \color{gray}- & \color{gray}- & \color{gray}- \\

                    \color{gray}Qwen3-VL-235B & \color{gray}235B
                    & \color{gray}50.5 & \color{gray}- & \color{gray}- & \color{gray}- & \color{gray}- & \color{gray}- & \color{gray}-
                    & \color{gray}92.4 & \color{gray}- & \color{gray}- & \color{gray}- & \color{gray}- & \color{gray}- & \color{gray}- \\
                    
                    \midrule
                    
                    Holo2-4B & 4B 
                    & 32.5 & 22.4 & 31.7 & 33.2 & 26.5 & 33.1 & 29.9 
                    & 88.9 & 81.7 & 90.0 & 86.2 & 82.0 & 89.4 & \underline{88.6} \\
                    
                    GUI-Actor-7B & 7B 
                    & 28.6 & 19.9 & 24.5 & 29.1 & 23.7 & 29.3 & 28.9 
                    & 88.4 & \underline{84.3} & 89.5 & \underline{88.6} & 83.9 & 89.2 & 85.2 \\
                    \midrule
                    
                    \rowcolor{icmlblue}
                    \textbf{TT-4B-ScreenSpot} & \textbf{4B}
                    & 29.3 & 24.5 & 28.9 & 29.4 & 25.6 & 30.1 & 29.0
                    & 86.8 & 83.4 & 86.9 & 85.3 & 84.1 & 89.9 & 86.2 \\
                    
                    \rowcolor{icmlblue}
                    \textbf{TT-4B-ScreenSpot-Pro} & \textbf{4B}
                    & \textbf{42.5} & \textbf{38.2} & \textbf{42.6} & \textbf{41.9} & \textbf{40.7} & \textbf{43.0} & \textbf{41.8}
                    & \underline{89.5} & \textbf{85.3} & \underline{90.1} & \textbf{90.6} & \textbf{87.4} & \underline{90.3} & 88.2 \\
                    \bottomrule
                \end{NiceTabular*}
            \end{sc}
        \end{scriptsize}
    \end{center}
    \vskip -0.2in
\end{table*}

\paragraph{Stage II Training.} The model is further fine-tuned on processed real-world GUI screenshots using the synthesized oracle trajectories $\mathcal{D}_{traj}$. To prevent the model from prioritizing the lengthy reasoning generation ($r_t$) over the crucial action execution ($a_t$), we utilize an \textbf{Action-Weighted Loss}. We assign a higher weight $\lambda_{act}$ (e.g., $\lambda_{act}=20$) specifically to the action tokens:
\begin{equation}
    \mathcal{L}_{SFT} = - \sum_{t=1}^{T} w_t \cdot \log \pi_\theta(y_t | \tilde{I}_t, x, h_{t-1}),
\end{equation}
where $w_t = \lambda_{act}$ if $y_t \in \mathcal{V}_{tool}$, and $w_t = 1$ otherwise. After stage II training, the model gains the capability to perform GUI tasks on real-world screenshots.

\paragraph{Stage III Training.} GUI tasks and datasets vary significantly in difficulty. To create an effective curriculum, stage II training cannot introduce too difficult tasks to prevent overfitting. Therefore, we implement another stage of training with similar settings as stage II where we utilize more difficult datasets to further adjust the agent to understand more difficult scenarios.

\subsection{Discussion}
\vspace{-0.3em}
The \textbf{\name{}} paradigm offers distinct advantages over coordinate-based methods. By replacing ambiguous coordinates with explicit semantic actions, it significantly enhances robustness, while our semantic anchoring and curriculum learning strategies ensure high data efficiency. Furthermore, the multi-step nature of our approach supports test-time scaling capabilities.

\section{Experiments}
\label{sec:experiments}

\subsection{Experimental Setup}

\vspace{-0.3em}
\paragraph{Implementation Details.} We instantiate ToolTok (TT) using the recently released \textit{Qwen3-VL-4B-Instruct}~\cite{bai2025qwen3vltechnicalreport} as the foundation to capitalize on visual priors. To rigorously assess the influence of data distribution,  we develop two variants: \textit{TT-4B-ScreenSpot} and \textit{TT-4B-ScreenSpot-Pro}. The ``Pro'' variant incorporates the more challenging ScreenSpot-Pro~\cite{li2025screenspot} dataset, enabling a systematic investigation into the model’s robustness within complex GUI environments. To mitigate the imbalance between extensive reasoning chains and single-step action tokens, we employ an Action-Weighted Loss (Sec.~\ref{sec: training_stategy}) within a three-stage progressive pipeline optimized via AdamW~\cite{loshchilov2017decoupled}. Comprehensive training specifications are provided in Appendix~\ref{app:training_details}.

\vspace{-0.3em}
\paragraph{Baseline Models.} To demonstrate the superiority of our approach, we benchmark it against models of both comparable scale and larger parameters. We categorize baselines into two groups. First, we consider \textbf{Generalist VLMs} represented by the \textit{4B}, \textit{8B}, and SOTA \textit{235B (MoE)} variants of \textit{Qwen3-VL-Instruct}~\cite{bai2025qwen3vltechnicalreport}. These models represent the prevailing paradigm of native coordinate regression. Second, we evaluate \textbf{Specialized GUI Agents}, including \textit{Holo2-4B}~\cite{hai2025holo2modelfamily} and \textit{GUI-Actor-7B}~\cite{xu2025attention}. While \textit{Holo2-4B}~\cite{hai2025holo2modelfamily} is fine-tuned from the same base model, it adheres to the conventional coordinate regression paradigm. On the other hand \textit{GUI-Actor-7B}~\cite{xu2025attention} relies on direct token-to-patch attention alignment for action modeling.

\vspace{-0.3em}
\paragraph{Datasets \& Benchmarks.} We evaluate our approach across four datasets, categorized into \textbf{In-domain (ID)} and \textbf{Out-of-domain (OOD)} scenarios. For ID datasets, we utilize \textit{ScreenSpot}~\cite{cheng2024seeclick} and its enhanced version, \textit{ScreenSpot-Pro}~\cite{li2025screenspot}, to establish and refine the agent's core navigation capabilities, leveraging the latter's professional-grade applications and dense UI elements for fine-grained interaction. To further evaluate the zero-shot generalization , we benchmark on \textit{Mind2Web-Simplified} and \textit{ScreenSpot-v2}~\cite{wu2024atlas}. For \textit{Mind2Web}~\cite{deng2023mind2web}, we center-crop a $2000 \times 2000$ region around the target with spatial jitter to evaluate local discrimination in dense web layouts. \textit{ScreenSpot-v2}~\cite{wu2024atlas} is further incorporated to benchmark generalization against more diverse scenarios and complex instructions.

\begin{figure*}[ht]
  \begin{center}
    \centerline{\includegraphics[width=\textwidth]{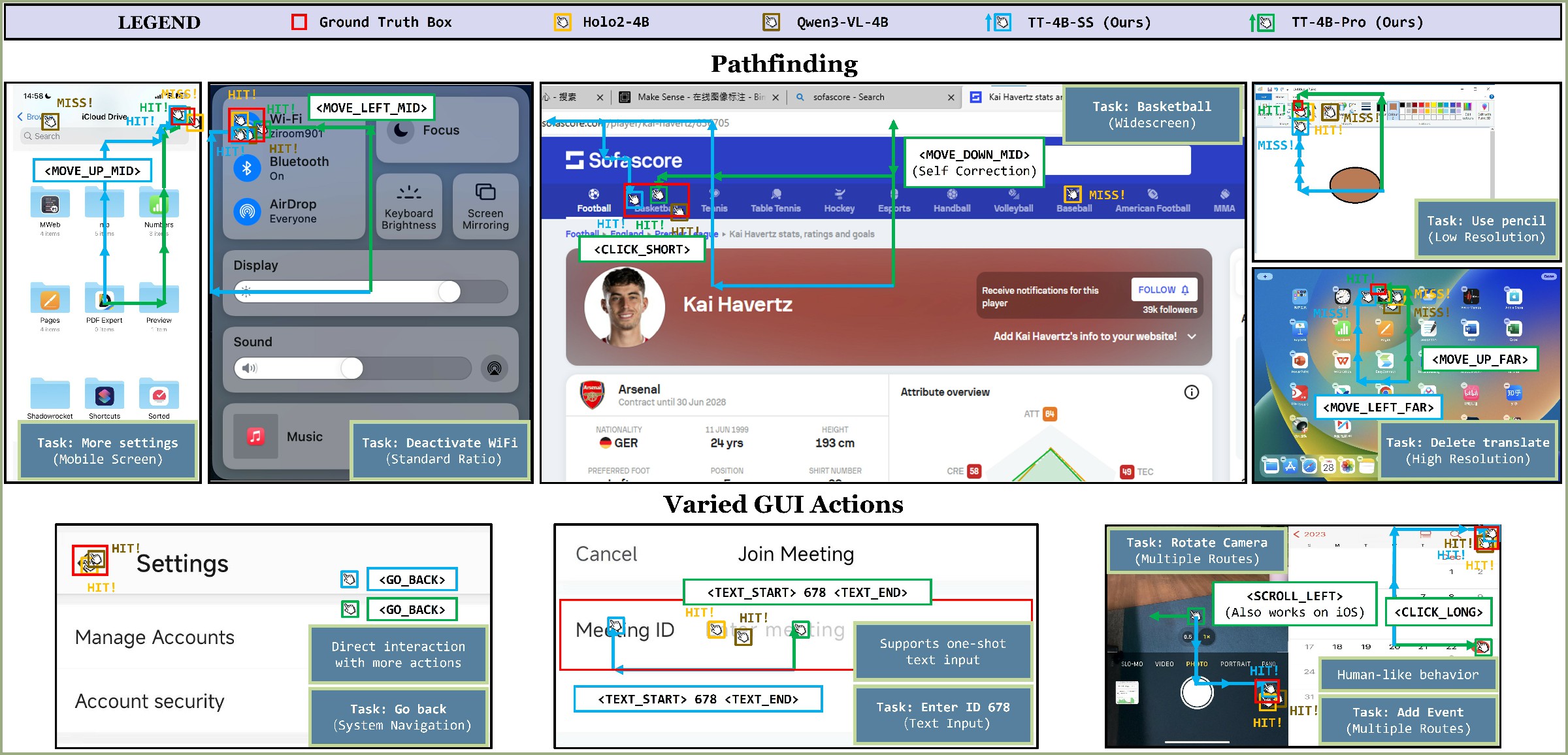}}
    \caption{
      Visualization for multiple experiment cases of different resolutions, ratios and usage scenarios. Our models generate intuitive and easy to understand paths while also supporting more varied and human-like actions. Row 1 samples show the superior semantic robustness of our models (self correction, shortcutting, etc.), while row 2 samples show emergent generalization capabilities where our models output valid but untrained actions.
    }
    \label{fig:vis1}
  \end{center}
  \vskip -0.3in
\end{figure*}

\paragraph{Evaluation Metrics.} Performance is primarily quantified via \textbf{Action Accuracy (Acc)}, which strictly requires both precise cursor positioning within the target bounding box and a correct interaction token match (\eg\texttt{<CLICK\_SHORT>}). Additionally, to assess \textbf{Robustness}, we evaluate performance consistency across varying input resolutions (e.g., Low, High) and aspect ratios (e.g., Mobile, Widescreen), measuring the model's ability to generalize beyond standard training configurations.

\subsection{Comparison Results.}

\vspace{-0.3em}
\paragraph{Benchmark Results} As shown in Tab.~\ref{tab:main_results}, \textit{TT-4B-ScreenSpot-Pro} outperforms both generalist baselines and specialized agents of comparable scale across all ID and OOD benchmarks in Action Accuracy (Acc.), especially for the more difficult \textit{ScreenSpot-Pro} and \textit{Mind2Web-Simplified}. This underscores \textbf{the efficacy of our progressive visual pathfinding fine-tuning paradigm} over the conventional coordinate regression approach. Notably, our \textit{TT-4B} delivers performance comparable to \textit{Qwen3-VL-235B}~\cite{bai2025qwen3vltechnicalreport}, demonstrating the robust potential of our proposed approach. To further isolate the contribution of our method, we evaluate the massive generalist model in a zero-shot pathfinding setting (\textit{Qwen3-VL-235B-PF}); its inferior performance confirms that the pathfinding capability is not intrinsic to the backbone but a direct result of our method. Finally, our substantial lead over \textit{GUI-Actor-7B}~\cite{wu2025gui} validates \textbf{the superiority of action discretization} over the token-to-patch attention alignment paradigm for precise GUI agency. Detailed prompts for the experiments are provided in Appendix.~\ref{app:prompts} for reproducibility.

\paragraph{Qualitative Analysis.} Fig. \ref{fig:vis1} visualizes inference trajectories across diverse input configurations, including varying resolutions, aspect ratios, and usage scenarios. As illustrated, \name{} generates intuitive, coarse-to-fine search paths that closely \textbf{align with human navigation patterns}. In contrast to visual grounding baselines, which are prone to severe semantic misalignment (e.g., confusing ``baseball" with ``basketball" due to visual-semantic ambiguity in Holo2-4B~\cite{hai2025holo2modelfamily}), our model exhibits superior \textbf{semantic robustness}. Even in instances of suboptimal precision, \name{} typically maintains valid spatial localization near the target region rather than drifting to semantically unrelated elements. Furthermore, the model demonstrates \textbf{emergent generalization capabilities}: it can synthesize correct, out-of-distribution action sequences for unseen instructions (Varied GUI Actions in Fig.~\ref{fig:vis1}), thereby validating the effectiveness of the underlying semantic anchoring and reasoning mechanisms.

\begin{table}[ht]
    \caption{\textbf{Data Efficiency Analysis}. Our method achieves superior performance with significantly less data. Data Efficiency (DE) is calculated as $\text{Avg. Acc}~/~\text{Data Scale (1M)}$ on \textit{ScreenSpot}.}
    \label{tab:data_efficiency}
    \begin{center}
        \begin{small}
            \begin{sc}
                \setlength{\tabcolsep}{2.5pt}
                \resizebox{0.4\textwidth}{!}{ 
                \begin{tabular}{lccccc}
                    \toprule
                    \raisebox{-1.5ex}[0pt][0pt]{\textbf{Model}} & \multicolumn{3}{c}{\textbf{Accuracy (\%)}} & \raisebox{-0.7ex}[0pt][0pt]{\textbf{Data}} & \raisebox{-1.5ex}[0pt][0pt]{\textbf{DE}$\uparrow$} \\
                    \cmidrule(lr){2-4}
                     & SS & Pro & M2W & \raisebox{+0.7ex}[0pt][0pt]{\textbf{Scale}} &  \\
                    \midrule
                    Holo2-4B      & 90.8 & 57.9 & 32.5 & - & - \\
                    SeeClick      & 53.4 & 1.5 & - & 1M & 53.4 \\
                    GUI-Actor-7B     & 85.9 & 44.6 & 23.7 & 1M & 85.9 \\
                    \midrule
                    \textbf{Ours} & \textbf{91.8} & \textbf{61.1} & \textbf{42.5} & \textbf{2K} & \textbf{43800} ($509\times$) \\
                    \bottomrule
                \end{tabular}}
            \end{sc}
        \end{small}
    \end{center}
    \vskip -0.3in
\end{table}

\begin{figure}[ht]
  \begin{center}
    \centerline{\includegraphics[width=\linewidth]{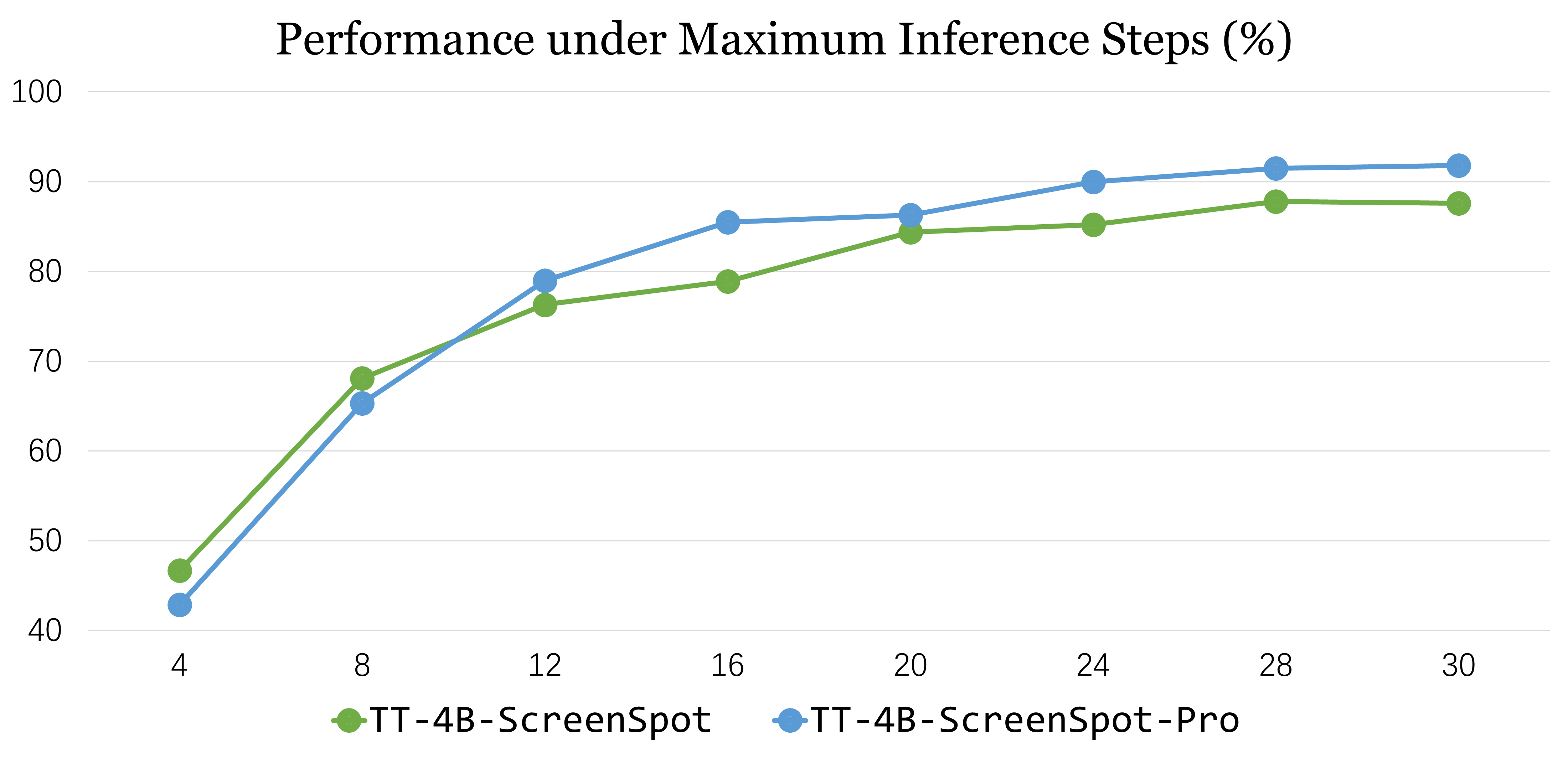}}
    \caption{
      Test-time scaling results of our models. As the maximum allowed step counts increases, the performance of our models significantly improves.
    }
    \label{fig:tts}
  \end{center}
  \vskip -0.3in
\end{figure}

\vspace{-0.3em}
\paragraph{Data Efficiency Analysis.} \tab~\ref{tab:data_efficiency} highlights the training cost-effectiveness. Unlike coordinate-based baselines requiring massive supervision ($\sim$1M samples), ToolTok achieves a Data Efficiency (DE) score of 43800, gaining more than $\bf{500\times}$ improvement using merely $\sim$2K real-world samples. This efficiency derives from semantic anchoring and curriculum learning, which effectively align discrete tools with visual priors via synthetic warm-up. Furthermore, the underlying CoT mechanism enables \textbf{test-time scaling}~\cite{snell2024scaling}, as shown in \fig \ref{fig:tts}, allocating more inference steps consistently yields performance gains, validating the benefit of computational scaling during inference.

\vspace{-0.3em}
\subsection{Robustness Experiments}
\vspace{-0.3em}
We evaluate resolution and ratio robustness across in and out of domain benchmarks shown in Tab.~\ref{tab:main_results}. Moreover, to validate the knowledge retention of our paradigm, which ensures fine-tuning does not compromise pre-trained capabilities, we benchmark on \textit{SimpleVQA}~\cite{cheng2025simplevqa} for visual factuality and \textit{MIA-Bench}~\cite{qian2024mia} for multimodal instruction following.

\paragraph{Resolution Robustness.} We resize samples (retaining the original ratio) until the longer side reaches the desired length to form the test set. Results are shown in \fig \ref{fig:robustness}. Our model is more consistent and achieves significantly better performance in low resolution and high resolution scenarios, which indicates better robustness.

\begin{table}[ht]
    \caption{Performance comparison on multimodal benchmarks. We report the accuracy ($\%$) on \textit{SimpleVQA} and \textit{MIA-Bench}. \textbf{TT-4B} variants denote our proposed models. The best results are highlighted in \textbf{bold} while the second best results are \underline{underlined}.}
    \label{tab:retention}
    \begin{center}
        \begin{small}
            \begin{sc}
                \setlength{\tabcolsep}{4pt}
                \resizebox{0.4\textwidth}{!}{ 
                \begin{tabular}{lcc}
                    \toprule
                    \raisebox{-1.5ex}[0pt][0pt]{\textbf{Model}} & \multicolumn{2}{c}{\textbf{Accuracy (\%)}} \\
                    \cmidrule(lr){2-3}
                                & SimpleVQA & MIA-Bench \\
                    \midrule
                    \textit{Original} \\
                    Qwen3-VL-4B-Instruct & \textbf{48.0} & \textbf{89.7} \\
                    \midrule
                    \textit{Fine-Tuned} \\
                    Holo2-4B    & 28.9 & 72.4 \\
                    TT-4B-ScreenSpot    & \underline{46.2} & \underline{88.9} \\
                    TT-4B-ScreenSpot-Pro   & 44.7 & 85.6 \\
                    \bottomrule
                \end{tabular}}
            \end{sc}
        \end{small}
    \end{center}
    \vskip -0.1in
\end{table}

\begin{figure}[ht]
  \begin{center}
    \centerline{\includegraphics[width=\linewidth]{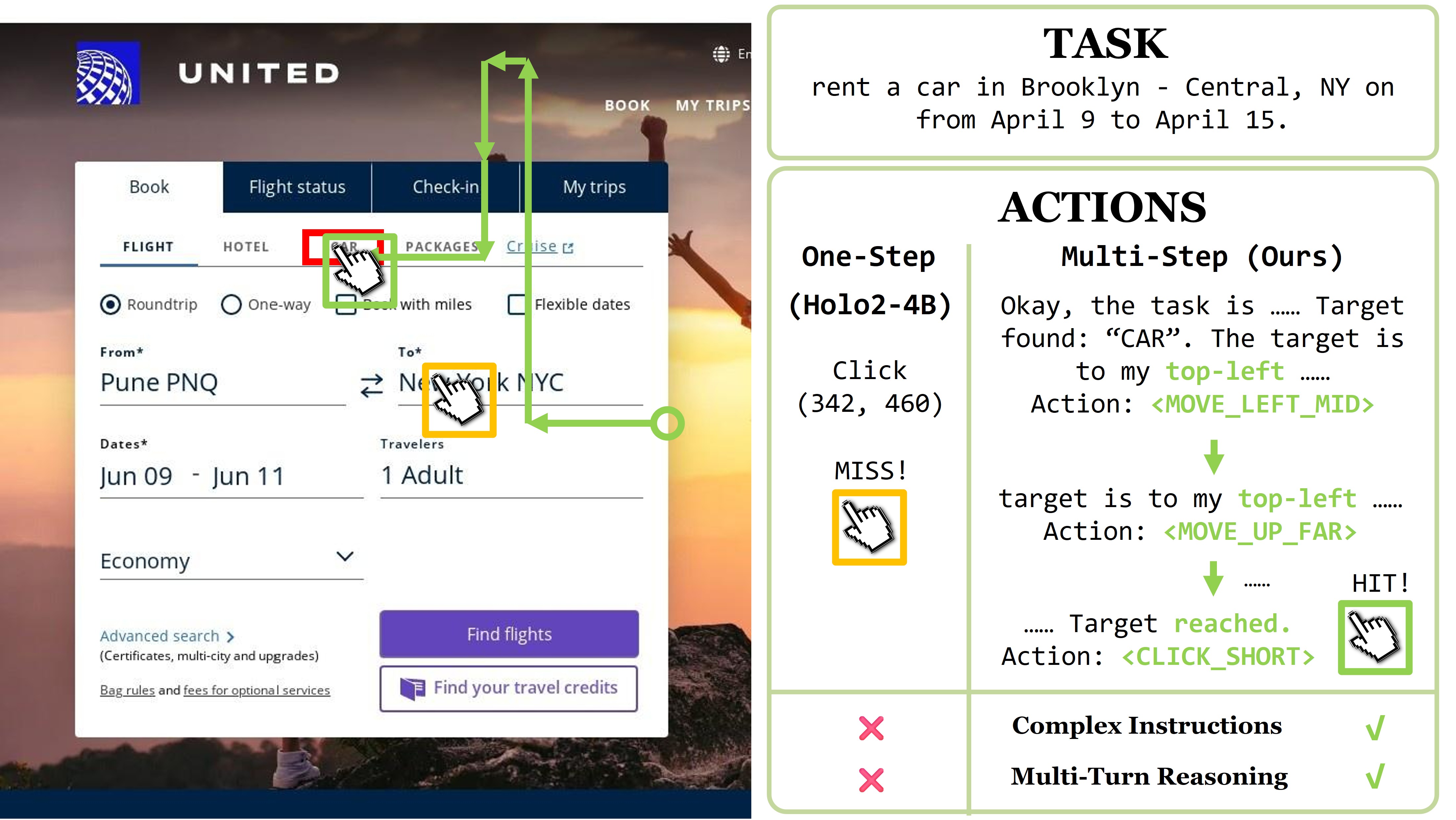}}
    \caption{
      Visualization of a sample from \textbf{Mind2Web-Simplified}. Our model correctly understands that the current step requires selecting the ``car'' tab while the baseline (\textit{Holo2-4B}) is disrupted by the long instruction and clicks on ``New York'' (Possibly due to ``NY'' being present in the instruction).
    }
    \label{fig:vis2}
  \end{center}
  \vskip -0.25in
\end{figure}

\vspace{-0.3em}
\paragraph{Ratio Robustness.} We select samples with desired ratios from \textit{ScreenSpot}~\cite{cheng2024seeclick} to form the test set. Results are in \fig \ref{fig:robustness}. Our model stays consistent and achieves superior performance in both mobile and widescreen ratio settings. This too indicates better robustness.

\paragraph{Knowledge Retention.} While fine-tuning often compromises pre-trained capabilities~\cite{zhai2023investigating}, \tab \ref{tab:retention} demonstrates that ToolTok preserves the backbone's general knowledge significantly better than baselines. This retention \textbf{facilitates superior comprehension of complex instructions} (e.g., in \textit{Mind2Web}), as illustrated in \fig \ref{fig:vis2}, where our model correctly parses the directive while \textit{Holo2-4B} fails due to semantic ambiguity.

\subsection{Ablation Study}
\vspace{-0.3em}
\begin{table}[ht]
    \vspace{-2pt}
    \caption{Ablation study on token initialization strategies. We report the execution accuracy (\%) on \textit{ScreenSpot}, \textit{ScreenSpot-Pro}, and \textit{Mind2Web-Simplified} (M2W-S). \textit{Ours} (\textit{TT-4B-ScreenSpot}) denotes our proposed method with the final initialization strategy. The best results are highlighted in \textbf{bold}.}
    \label{tab:anchoring}
    \begin{center}
        \begin{small}
            \begin{sc}
                \setlength{\tabcolsep}{11pt}
                \resizebox{0.4\textwidth}{!}{ 
                \begin{tabular}{lccc}
                    \toprule
                    \raisebox{-1.5ex}[0pt][0pt]{\textbf{Initialization}} & \multicolumn{3}{c}{\textbf{Accuracy (\%)}} \\
                    \cmidrule(lr){2-4}
                                & SS & SS-Pro & M2W-S \\
                    \midrule
                    zero-init      & 55.2 & 23.4 & 15.3 \\
                    rand-init & 56.4 & 22.1 & 16.5 \\
                    avg-init    & 68.5 & 32.9 & 19.0 \\
                    \midrule
                    \textit{Ours} & \textbf{87.6} & \textbf{50.2} & \textbf{29.3} \\
                    \bottomrule
                \end{tabular}}
            \end{sc}
        \end{small}
    \end{center}
    \vskip -0.2in
\end{table}

\begin{figure}[ht]
  \begin{center}
    \centerline{\includegraphics[width=\linewidth]{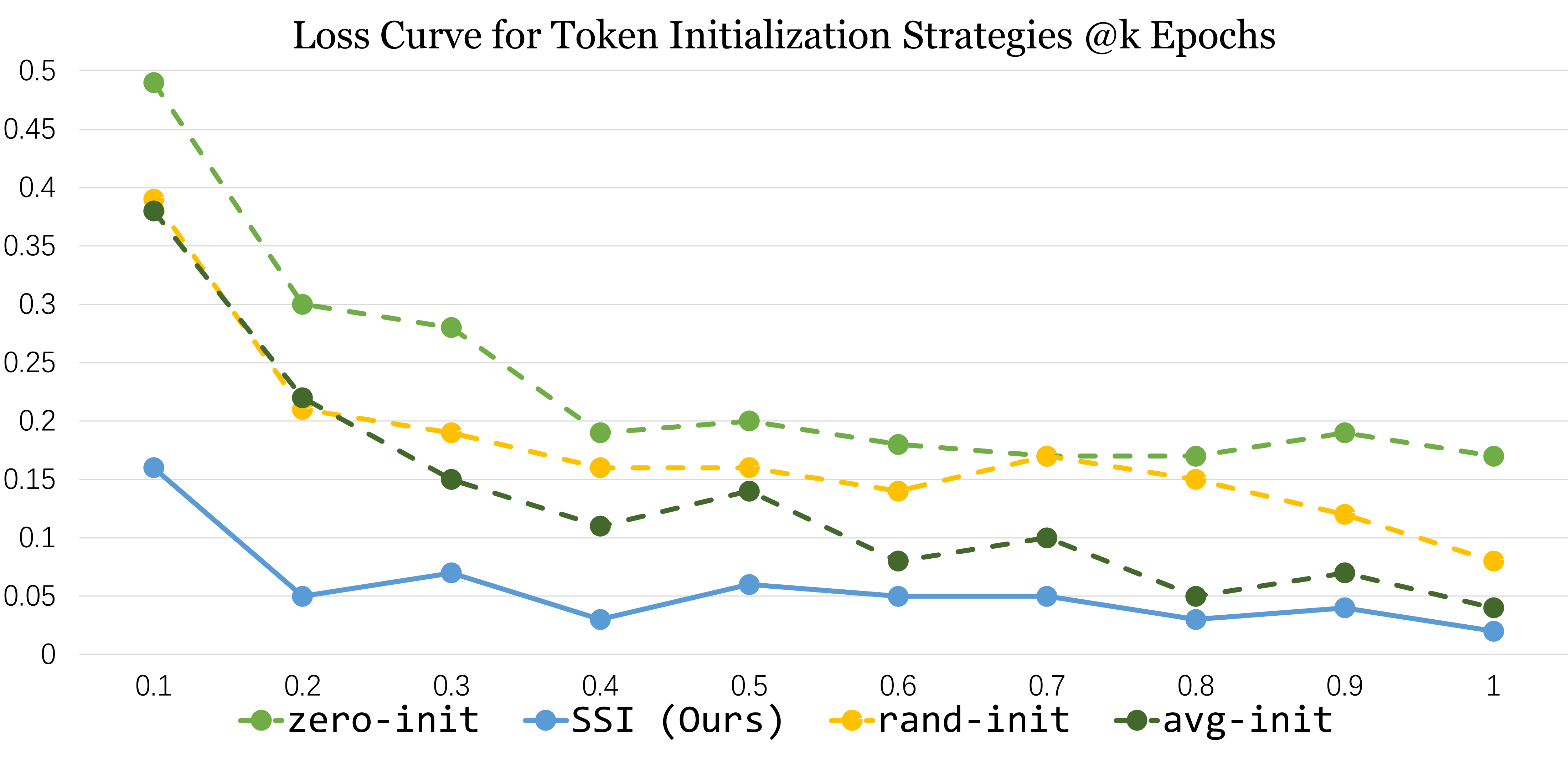}}
    \caption{
      Loss curves for models using different token initialization strategies. Among all methods that include new tokens, our method, \textbf{SSI} has the fastest convergence speed.
    }
    \label{fig:init}
  \end{center}
  \vskip -0.3in
\end{figure}

\vspace{-0.3em}
\paragraph{Effects of Semantic Anchoring.} To isolate the efficacy of semantic anchoring in token embedding initialization, we benchmark our Spherical Semantic Initialization (SSI) against four alternative strategies. We establish lower-bound baselines using \textit{Zero Initialization (zero-init)}, which renders new tokens initially indistinguishable. For a standard reference, we employ \textit{Random Initialization (rand-init)}~\cite{lester2021power}, the prevailing method for embedding expansion. Finally, to validate the necessity of spherical projection, we compare against \textit{Simple Average Anchoring (avg-init)}; while this variant utilizes semantic anchors, it computes an arithmetic mean without norm constraints. As evidenced by the training dynamics in \fig \ref{fig:init} and accuracy in \tab \ref{tab:anchoring}, SSI not only significantly accelerates convergence but also yields superior final performance.
\begin{table}[ht]
    \caption{Ablation study on different training curriculums. We report the execution accuracy (\%) on \textit{ScreenSpot}, \textit{ScreenSpot-Pro}, and \textit{Mind2Web-Simplified} (M2W-S). The best results are highlighted in \textbf{bold} and the second best are \underline{underlined}.}
    \label{tab:training}
    \begin{center}
        \begin{small}
            \begin{sc}
                \setlength{\tabcolsep}{5pt}
                \resizebox{0.4\textwidth}{!}{
                \begin{tabular}{lccc}
                    \toprule
                    \raisebox{-1.5ex}[0pt][0pt]{\textbf{Setting}} & \multicolumn{3}{c}{\textbf{Accuracy (\%)}} \\
                    \cmidrule(lr){2-4}
                    & ScreenSpot & ScreenSpot-Pro & M2W-S \\
                    \midrule
                    Only Pro         & 72.0 & 48.5 & 22.8 \\
                    SS mix Pro  & 84.1 & \underline{52.9} & 24.4 \\
                    Only SS     & \underline{87.6} & 50.2 & \underline{29.3} \\
                    SS + Pro   & \textbf{91.8} & \textbf{61.1} & \textbf{42.5} \\
                    \bottomrule
                \end{tabular}
                }
            \end{sc}
        \end{small}
    \end{center}
    \vskip -0.2in
\end{table}

\paragraph{Effects of Curriculum Learning.} To validate our learning curriculum, we compare our progressive strategy against static baselines. We contrast our staged training—transitioning from the easier \textit{ScreenSpot} (SS) to the complex \textit{ScreenSpot-Pro} (Pro)—with three configurations: single-stage fine-tuning (\textbf{SS Only}, \textbf{Pro Only}) and a joint mixture strategy (\textbf{SS mix Pro}). As detailed in \tab \ref{tab:training}, the staged curriculum (\textbf{SS + Pro}) consistently outperforms both single-stage and mixed baselines across all benchmarks. This confirms that structured progression from easy to hard tasks, rather than mere data aggregation, is essential for robust GUI pathfinding.

\section{Conclusion}
\vspace{-0.3em}
In this paper, we propose \textbf{\name{}}, a new paradigm for GUI agents that shifts from visual grounding to visual pathfinding, discretizing GUI operations into tool tokens to enable robust multi-step reasoning. Utilizing semantic anchoring and our $3$ stage training curriculum, we achieve SOTA in multiple datasets among 4B models with \textbf{less than 1\%} of non-synthesized training data compared to previous approaches. \textbf{\name{}} is a promising paradigm shift for future GUI agent works, while \textbf{semantic anchoring} can be effectively transferred to other special token alignment scenarios. Overall, \textbf{\name{}} is a firm step towards more \textbf{human-like} agentic frameworks.


\section*{Impact Statement}

This paper presents \textbf{\name{}}, a method for developing data efficient and generalizable GUI agents. Our work has the potential to positively impact society by automating repetitive digital tasks and improving accessibility for users with physical impairments, enabling them to interact with computers more easily through natural language.

\bibliography{main}
\bibliographystyle{icml2026}

\newpage
\appendix
\onecolumn
\section{Related Works}
\label{app:related_works}

\vspace{-0.3em}
\paragraph{Vision-Language Models for GUI Agents.} While traditional agents relied on document object model (DOM) parsing \cite{lai2024autowebglm, zhang2025webpilot, gur2023real}, limiting them to web environments, recent Vision-Language Models (VLMs) have enabled pixel-based interaction. Current approaches, including generalist \cite{bai2025qwen3vltechnicalreport, guo2025seed15vltechnicalreport, vteam2026glm45vglm41vthinkingversatilemultimodal} and specialized models \cite{cheng2024seeclick, liu2024autoglm, qin2025ui, hai2025holo2modelfamily,lee2023pix2struct, wang2024mobile, gou2024navigating}, predominantly adopt a \textbf{visual grounding} paradigm based on \textbf{coordinates}. However, reliance on fixed coordinate schemes \cite{nguyen2025gui} hampers robustness and generalization. Although coordinate-free alternatives like TAG \cite{wu2025gui} and GUI-Actor \cite{xu2025attention} have emerged, they remain limited to simple actions.

\paragraph{GUI Datasets \& Benchmarks.} Despite the abundance of multimodal GUI datasets and benchmarks \cite{li2025screenspot, cheng2024seeclick, wu2024atlas, deng2023mind2web, xie2024osworld, he2024webvoyager, lu2025guiodyssey}, a significant gap remains. High-quality expert-annotated datasets are typically limited in scale, while larger, auto-generated alternatives often suffer from noise and poor captioning. Critically, the vast majority provide only ground truth bounding boxes without cursor movement history, creating a data challenge for developing GUI agents that employ a multi-step pathfinding paradigm.

\paragraph{Action \& Tool Representation.} Representing actions or tools as discrete, learnable tokens has emerged as a powerful paradigm in NLP and Robotics. \textbf{ToolkenGPT} \cite{hao2024toolkengptaugmentingfrozenlanguage} pioneers the concept of ``toolkens''--learnable embeddings that enable frozen LLMs to master massive tools without fine-tuning parameters. Similarly, \textbf{Toolscaler} \cite{su2025toolscaler} optimizes generative tool calling through structure-aware tokenization, while works in the field of robotics \cite{zitkovich2023rt, bousmalis2023robocat} demonstrate the efficacy of discretizing continuous control signals into tokens for generalized robotic manipulation.  However, the ``cold start'' problem for tool token alignment is significant due to randomized initial embeddings. Therefore, tool tokenization is yet to be implemented in the field of GUI agents as a result of limited high-quality data.

\section{Design Details}

\subsection{Action Tokens}
\label{app:token_definition}

\paragraph{Token Definition.} We categorize the tool tokens into four types. We define each token as: 

\begin{itemize}
\item \textbf{$\texttt{<MOVE\_UP\_FAR>}$:} Move the cursor up by $500$px.
\item \textbf{$\texttt{<MOVE\_UP\_MID>}$:} Move the cursor up by $150$px.
\item \textbf{$\texttt{<MOVE\_UP\_CLO>}$:} Move the cursor up by $30$px.
\item \textbf{$\texttt{<MOVE\_DOWN\_FAR>}$:} Move the cursor down by $500$px.
\item \textbf{$\texttt{<MOVE\_DOWN\_MID>}$:} Move the cursor down by $150$px.
\item \textbf{$\texttt{<MOVE\_DOWN\_CLO>}$:} Move the cursor down by $30$px.
\item \textbf{$\texttt{<MOVE\_LEFT\_FAR>}$:} Move the cursor left by $500$px.
\item \textbf{$\texttt{<MOVE\_LEFT\_MID>}$:} Move the cursor left by $150$px.
\item \textbf{$\texttt{<MOVE\_LEFT\_CLO>}$:} Move the cursor left by $30$px.
\item \textbf{$\texttt{<MOVE\_RIGHT\_FAR>}$:} Move the cursor right by $500$px.
\item \textbf{$\texttt{<MOVE\_RIGHT\_MID>}$:} Move the cursor right by $150$px.
\item \textbf{$\texttt{<MOVE\_RIGHT\_CLO>}$:} Move the cursor right by $30$px.
\item \textbf{$\texttt{<GO\_HOME>}$:} Return to the home tab or desktop.
\item \textbf{$\texttt{<GO\_BACK>}$:} Return to the last tab or access level.
\item \textbf{$\texttt{<CLICK\_SHORT>}$:} Perform a short mouse click at the current position.
\item \textbf{$\texttt{<CLICK\_LONG>}$:} Perform a long click (or a hold) at the current position.
\item \textbf{$\texttt{<SCROLL\_UP>}$:} Scroll or swipe the screen up.
\item \textbf{$\texttt{<SCROLL\_DOWN>}$:} Scroll or swipe the screen down.
\item \textbf{$\texttt{<SCROLL\_LEFT>}$:} Scroll or swipe the screen left.
\item \textbf{$\texttt{<SCROLL\_RIGHT>}$:} Scroll or swipe the screen right.
\item \textbf{$\texttt{<TEXT\_START>}$:} Starting indicator of text input.
\item \textbf{$\texttt{<TEXT\_END>}$:} Ending indicator of text input. Usage: $\texttt{<TEXT\_START>}$ Input text $\texttt{<TEXT\_END>}$.
\end{itemize}

\paragraph{Semantic Anchor Words for Each Token.} We define a list of words that act as semantic anchors for each action token, provided in Listing \ref{lst:anchors}.

\begin{lstlisting}[
    basicstyle=\footnotesize\ttfamily,
    breaklines=true,
    frame=single,
    columns=fullflexible,
    caption={Semantic Anchor Words of Action Tokens},
    label={lst:anchors}
]
    "<MOVE_UP_FAR>":   ["move", "cursor", "up", "top", "far", "jump", "leap"],
    "<MOVE_UP_MID>":   ["move", "cursor", "up", "medium", "shift"],
    "<MOVE_UP_CLO>":   ["move", "cursor", "up", "near", "tiny", "nudge", "slight"],

    "<MOVE_DOWN_FAR>": ["move", "cursor", "down", "bottom", "far", "jump", "leap"],
    "<MOVE_DOWN_MID>": ["move", "cursor", "down", "medium", "shift"],
    "<MOVE_DOWN_CLO>": ["move", "cursor", "down", "near", "tiny", "nudge", "slight"],

    "<MOVE_LEFT_FAR>": ["move", "cursor", "left", "west", "far", "jump", "leap"],
    "<MOVE_LEFT_MID>": ["move", "cursor", "left", "medium", "shift"],
    "<MOVE_LEFT_CLO>": ["move", "cursor", "left", "near", "tiny", "nudge", "slight"],

    "<MOVE_RIGHT_FAR>": ["move", "cursor", "right", "east", "far", "jump", "leap"],
    "<MOVE_RIGHT_MID>": ["move", "cursor", "right", "medium", "shift"],
    "<MOVE_RIGHT_CLO>": ["move", "cursor", "right", "near", "tiny", "nudge", "slight"],

    "<CLICK_SHORT>": ["click", "tap", "select", "press", "touch", "brief", "mouse"],
    "<CLICK_LONG>":  ["click", "hold", "press", "long", "keep", "sustain"],

    "<GO_BACK>": ["navigate", "back", "return", "previous", "history", "reverse", "phone", "mobile"],
    "<GO_HOME>": ["navigate", "home", "main", "dashboard", "desktop", "launcher", "phone", "mobile"],

    "<SCROLL_UP>":    ["scroll", "swipe", "pan", "up", "top", "content"],
    "<SCROLL_DOWN>":  ["scroll", "swipe", "pan", "down", "bottom", "content"],
    "<SCROLL_LEFT>":  ["scroll", "swipe", "pan", "left", "west", "content"],
    "<SCROLL_RIGHT>": ["scroll", "swipe", "pan", "right", "east", "content"],

    "<TEXT_START>": ["type", "write", "input", "keyboard", "start", "focus", "activate"],
    "<TEXT_END>":   ["type", "finish", "enter", "submit", "confirm", "return"],
\end{lstlisting}

\subsection{Data Synthesis}
\label{app:data_synth}

We synthesize $3$ types of data for stage I training:

\begin{itemize}[leftmargin=1em, itemsep=0pt, topsep=0pt]
    \item \textbf{Token Definition QA:} We generate Question-Answer pairs (e.g., ``What does \texttt{<MOVE\_UP\_FAR>} mean?'') to explicitly teach the agent the functional definitions of each action token.
    \item \textbf{Text-Guided Tool Selection:} To bridge natural language instructions with actions, we generate samples where the agent must select the correct tool token based solely on textual descriptions of displacement (e.g., ``Move the cursor upwards by a large distance'' $\rightarrow$ \texttt{<MOVE\_UP\_FAR>}).
    \item \textbf{Simplified Visual Pathfinding:} There is a large gap between language understanding and complex spatial reasoning. To train the model to connect tool tokens with visual cues, we generate simplified canvases with random GUI backgrounds containing a rendered cursor, a target bounding box, and multiple \textit{distractor targets} (fake elements). The agent is trained to navigate to the correct target using the optimal sequence of actions. This forces the model to learn the spatial dynamics of the hierarchical action space (e.g., chaining \texttt{FAR} and \texttt{CLO} moves) without the interference of complex real-world task semantics.
\end{itemize}

\begin{figure}[ht]
  \vskip 0.2in
  \begin{center}
    \centerline{\includegraphics[width=\linewidth]{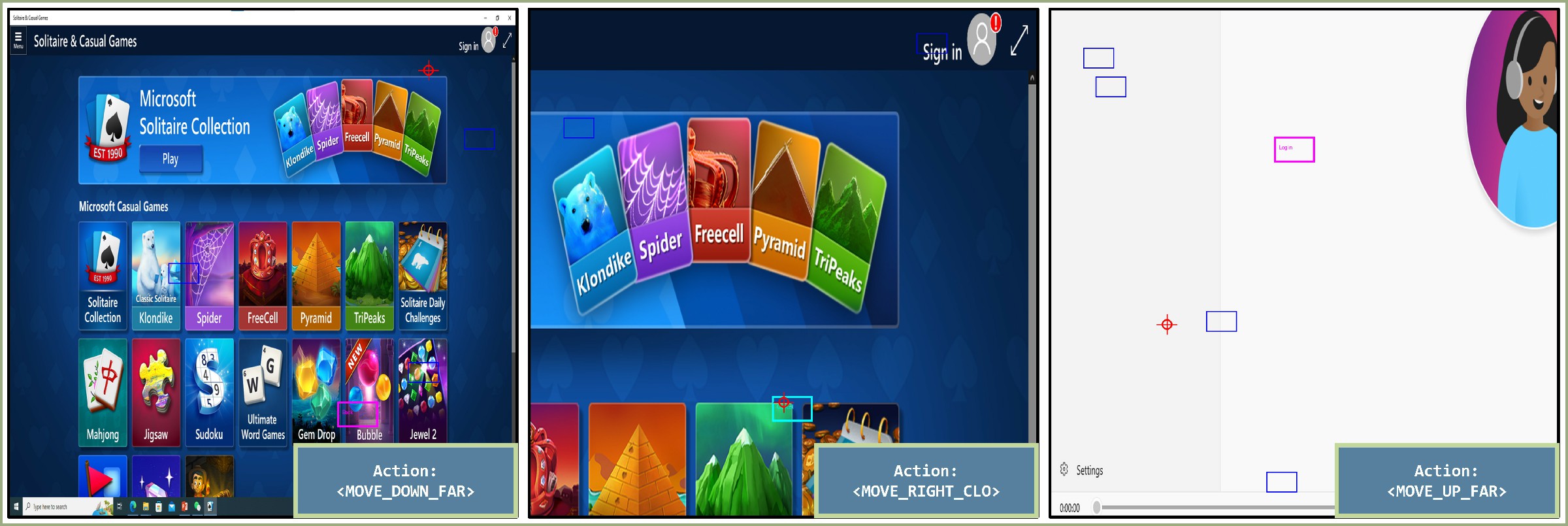}}
    \caption{
      Synthesized data examples that feature a red target as the cursor, some noise rectangles and a target rectangle with captioning.
    }
    \label{fig:synth_data}
  \end{center}
  \vskip -0.2in
\end{figure}

\fig \ref{fig:synth_data} provides some examples of synthesized \textbf{Simple visual pathfinding}. We construct a total of $5,000$ synthetic data points, with $1000$ \textit{token definition QA} problems, $2000$ \textit{text-guided selection} problems, and $2000$ \textit{simplified visual pathfinding} problems.

\subsection{CoT Construction}
\label{app:cot}

As implemented in our data augmentation pipeline, our CoT decomposes the decision process into three detailed logical stages:

\begin{itemize}[leftmargin=1em, itemsep=0pt, topsep=0pt]
    \item \textbf{Visual Localization:} The agent first anchors its state by identifying the absolute grid region of the cursor (e.g., ``Cursor is currently in the \textbf{Top-Left} region'') and the relative direction of the target (e.g., ``Target is to the \textbf{Bottom-Right}''). For targets within the same grid region, it further refines this with approximated relative coordinates to enhance local precision.
    \item \textbf{Axis Prioritization:} To determine the optimal movement direction, the agent explicitly compares the horizontal ($dx$) and vertical ($dy$) distances to the target. It prioritizes the dominant axis with the larger gap (e.g., ``Currently the \textbf{RIGHT} direction is the farthest away''), ensuring efficient rectilinear navigation.
    \item \textbf{Action Scaling:} Finally, the agent maps the quantitative pixel distance to the qualitative semantics of our tool tokens. It justifies the choice based on predefined thresholds (e.g., ``There is a significant gap. I need a large jump $\rightarrow$ use \texttt{FAR}'').
\end{itemize}

This structured reasoning provides a dense supervision signal that bridges the gap between high-level visual perception and low-level discrete control. 

\section{Training Details}
\label{app:training_details}
We build our models upon the \textit{Qwen3-VL-4B-Instruct} backbone~\cite{bai2025qwen3vltechnicalreport}. To adapt the model for GUI agency while preserving its general multi-modal capabilities, we freeze the vision encoder to retain the pre-trained visual representations and only fine-tune the LLM backbone and the embedding layers (including the newly initialized tool tokens). The backbone employs weight tying between $embedding$ and $lm\_head$ layers. We untie these layers before training to better adjust $lm\_head$ to output new tool tokens. We implement the training using the PyTorch framework. The model is trained using the AdamW~\cite{loshchilov2017decoupled} optimizer with a cosine learning rate schedule. The specific learning rate schedules for each stage are detailed in Tab.~\ref{tab:training_details} All experiments are conducted on a cluster of 8 NVIDIA A6000Ada GPUs.
\label{training_details}
\begin{table}[htbp]
    \centering
    \caption{Curriculum training details.}
    \begin{adjustbox}{max width=0.95\textwidth} 
\begin{tabular}{cccc}
\toprule
Setting & Stage I & Stage II &  Stage III \\
\midrule
LR.           & 1.5e-5 & 1e-5 & 5e-6 \\
Batch Size    & 8 & 8 & 8 \\
Training Epoch & 1 & 3 & 2 \\
Training Step & 580 & 360 & 320 \\
\bottomrule
\end{tabular}
\end{adjustbox}
    \label{tab:training_details}
\end{table}

Stage I is trained on our synthesized data. After finishing stage I, the model enters stage II. Stage II training is performed on \textit{ScreenSpot}, with about $1000$ samples used as training set. \textit{TT-4B-ScreenSpot} is the resulting model after stage II training. We take \textit{TT-4B-ScreenSpot} for further training. Stage III training utilizes \textit{ScreenSpot-Pro}, with about $1200$ samples used as training set. \textit{TT-4B-ScreenSpot-Pro} is the resulting model after stage III training.

For the training dataset, all in-domain data are partitioned into a strict 8:2 ratio for training and testing to eliminate potential data leakage (We do not use validation set due to extremely limited data size). We sequentially incorporate 1,000 samples from \textit{ScreenSpot} during Stage II to establish foundational grounding capabilities, followed by 1,200 samples from \textit{ScreenSpot-Pro} in Stage III for refinement. The latter—characterized by higher resolutions and denser UI elements in professional-grade applications—serves as a rigorous benchmark for fine-grained interaction.

\section{Experiment Details}
\label{app:experiment_details}

\subsection{Experiment Setup}

\paragraph{Main Experiment.} We perform the main experiment on \textit{ScreenSpot}, \textit{ScreenSpot-Pro}, \textit{Mind2Web-Simplified} and \textit{ScreenSpot-v2}. The experiment sample size is provided in \tab \ref{tab:experiment_details}.

We perform experiments on reproduced baselines. We run inference locally for \textit{Qwen3-VL-4B-Instruct}, \textit{Holo2-4B}, and \textit{GUI-Actor-7B}, while calling official API from Qwen to reproduce \textit{Qwen3-VL-8B-Instruct} and \textit{Qwen3-VL-235B-A22B-Instruct} due to limited computing resource. Noticing that in the official \textit{Qwen3-VL} cookbook, all GUI grounding input requires to be normalized to $1000*1000$, we follow suit.

\begin{table}[htbp]
    \centering
    \caption{Main Experiment Data Size}
    \begin{adjustbox}{max width=0.95\textwidth} 
\begin{tabular}{ccccc}
\toprule
Setting & ScreenSpot & ScreenSpot-Pro & Mind2Web-Simplified & ScreenSpot-v2 \\
\midrule
Data Size           & 240 & 320 & 1000 & 1000 \\
\bottomrule
\end{tabular}
\end{adjustbox}
    \label{tab:experiment_details}
\end{table}

\paragraph{Resolution Robustness.} We conduct resolution robustness experiments on \textbf{resized samples}. That is, we resize original samples using $\texttt{PIL.Image.LANCZOS}$ to the designated size (while retaining original ratios) before conducting experiments. Notice that resizing is not a lossless operation, the results in resolution robustness experiments are usually slightly lower for all models than those in the original setting. 

\paragraph{Ratio Robustness.} We select samples of the desired width-height ratio ($W:H$) to conduct ratio robustness experiments under $3$ categories: Mobile ($[0.4:0.6]$), standard ($[0.9:1.1]$), and widescreen ($[1.9:2.1]$). Strictly, these ratios should be $0.5:1$, $1:1$, and $2:1$. However, as not enough perfect samples exist, we approximate the ratios to a small range given above. This partition method is not to be confused with sample tags (iOS, Android, MacOS, etc.) of the datasets as the only relevant parameter in our ratio robustness experiments is aspect ratio.

\subsection{Prompts}
\label{app:prompts}

\paragraph{System prompt for TT-4B models.} System prompt is used to guide training and inference for \textbf{TT-4B} models. This pre-defines the task scope, creates context and enhances performance. The full system prompt used for our models is provided in Listing \ref{lst:prompt_sys}:

\begin{lstlisting}[
    basicstyle=\footnotesize\ttfamily,
    breaklines=true,
    frame=single,
    columns=fullflexible,
    caption={System Prompt for TT-4B Models},
    label={lst:prompt_sys}
]
You are a helpful GUI Agent.
If prompted with [Action], You can use the following action tokens:

<MOVE_UP_FAR>
<MOVE_DOWN_FAR>
<MOVE_LEFT_FAR>
<MOVE_RIGHT_FAR>
<MOVE_UP_MID>
<MOVE_DOWN_MID>
<MOVE_LEFT_MID>
<MOVE_RIGHT_MID>
<MOVE_UP_CLO>
<MOVE_DOWN_CLO>
<MOVE_LEFT_CLO>
<MOVE_RIGHT_CLO>
<CLICK_SHORT>
<CLICK_LONG>
<TEXT_START> [text] <TEXT_END>
<SCROLL_UP>
<SCROLL_DOWN>
<SCROLL_LEFT>
<SCROLL_RIGHT>
<GO_BACK>
<GO_HOME>
<END_ACTION>

Before you try to do non action commands like <GO_BACK> or <TEXT_START>, you should revisit the image and **ensure** that it **absolutely** cannot be done by navigating and clicking something on the screen.

You must output your response in two clearly labeled sections:

Reasoning: [Step-by-step analysis of the screen content and instruction]
Action: [The specific Action Token to execute]
\end{lstlisting}

\paragraph{System prompt for Qwen3-VL based baselines.} System prompt is necessary for using \textit{Qwen3-VL} based GUI agents to perform GUI tasks. We follow the official cookbook of \textit{Qwen3-VL}, using the following prompt scheme provided in Listing \ref{lst:prompt_baseline} for all visual grounding based \textit{Qwen3-VL} variants.

\begin{lstlisting}[
    basicstyle=\footnotesize\ttfamily,
    breaklines=true,
    frame=single,
    columns=fullflexible,
    caption={System Prompt for Qwen3-VL Based Visual Grounding Agents},
    label={lst:prompt_baseline}
]
You are a helpful assistant.

You may call one or more functions to assist with the user query.

You are provided with function signatures within <tools></tools> XML tags:
<tools>
{"type": "function", "function": {"name": "computer_use", "description": "Use a mouse to interact with a computer.\n* The screen's resolution is {WIDTH}x{HEIGHT}.\n* Make sure to click any buttons, links, icons, etc with the cursor tip in the center of the element. Don't click boxes on their edges unless asked.\n* you can only use the left_click and mouse_move action to interact with the computer. if you can't find the element, you should terminate the task and report the failure.", "parameters": {"properties": {"action": {"description": "The action to perform. The available actions are:\n* `mouse_move`: Move the cursor to a specified (x, y) pixel coordinate on the screen.\n* `left_click`: Click the left mouse button with coordinate (x, y).\n* `terminate`: Terminate the current task and report its completion status.", "enum": ["mouse_move", "left_click"], "type": "string"}, "coordinate": {"description": "(x, y): The x (pixels from the left edge) and y (pixels from the top edge) coordinates to move the mouse to. Required only by `action=mouse_move` and `action=left_click`.", "type": "array"}, "status": {"description": "The status of the task. Required only by `action=terminate`.", "type": "string", "enum": ["success", "failure"]}}, "required": ["action"], "type": "object"}}}
</tools>

For each function call, return a json object with function name and arguments within <tool_call></tool_call> XML tags:
<tool_call>
{"name": <function-name>, "arguments": <args-json-object>}
</tool_call>
\end{lstlisting}

Additionally, we provide the system prompt for \textit{Qwen3-VL-235B-PF}, the baseline for zero-shot visual pathfinding in Listing \ref{lst:prompt_pf}.

\begin{lstlisting}[
    basicstyle=\footnotesize\ttfamily,
    breaklines=true,
    frame=single,
    columns=fullflexible,
    caption={System Prompt for Qwen3-VL-235B-PF},
    label={lst:prompt_pf}
]
You are an intelligent GUI Agent controlling a cursor.

The cursor is a red crosshair with a round and four lines. You must identify the location of the cursor.

Your goal is to achieve the user's instruction by outputting specific Action Tokens.
You must strictly follow the format and vocabulary below.

**1. AVAILABLE ACTION TOKENS & PHYSICS:**

**A. Movement (Cursor Navigation)**
*Select the move stride based on the estimated pixel distance between the Cursor and the Target.*
*Image Size Reference: The screen is processed as a square grid (e.g., 1000x1000).*

- **Long-Range Jumps (Stride: 500px)**
  *Use when the gap is significant (> 300px).*
  - `<MOVE_UP_FAR>`: Move Up 500px.
  - `<MOVE_DOWN_FAR>`: Move Down 500px.
  - `<MOVE_LEFT_FAR>`: Move Left 500px.
  - `<MOVE_RIGHT_FAR>`: Move Right 500px.

- **Standard Navigation (Stride: 150px)**
  *Use when the target is moderately away (100px - 300px).*
  - `<MOVE_UP_MID>`: Move Up 150px.
  - `<MOVE_DOWN_MID>`: Move Down 150px.
  - `<MOVE_LEFT_MID>`: Move Left 150px.
  - `<MOVE_RIGHT_MID>`: Move Right 150px.

- **Micro-Adjustments (Stride: 30px)**
  *Use when the target is very close (< 100px) but NOT hit (< 15px).*
  - `<MOVE_UP_CLO>`: Nudge Up 30px.
  - `<MOVE_DOWN_CLO>`: Nudge Down 30px.
  - `<MOVE_LEFT_CLO>`: Nudge Left 30px.
  - `<MOVE_RIGHT_CLO>`: Nudge Right 30px.

**B. Interaction (Execution)**
*Perform these ONLY when the cursor is over the target (Distance < 30px).*

- `<CLICK_SHORT>`: **Primary Action.** Click the element. 
  *Condition:* Cursor MUST be overlapping the target.
- `<CLICK_LONG>`: Long press/Hold (e.g., for context menus).
- `<TEXT_START> [content] <TEXT_END>`: Type text. 
  *Condition:* Cursor must be over the input field (or field already active).
- `<GO_BACK>`: Return to previous page. (No cursor position required).
- `<GO_HOME>`: Return to system home. (No cursor position required).

**C. Termination**
- `<END_ACTION>`: Task fully complete.

---

**2. REASONING PROCESS (CHAIN OF THOUGHT):**

You must "think" strictly following this spatial analysis logic before acting:

1.  **Grid Localization:** Identify which 3x3 grid region (Top-Left, Center, Bottom-Right, etc.) the Cursor and Target are in.
2.  **Coordinate Estimation:** Estimate the **Relative Coordinates** [0.0 - 1.0] for both Cursor and Target.
    * *Format:* "More specifically, the cursor is at about [0.x, 0.y] and the target is at about [0.x, 0.y] (relative coordinates)..."
3.  **Direction Analysis:** Determine the relative direction (e.g., "The target is to the right of the cursor").
4.  **Axis Prioritization:** Compare the horizontal (X) and vertical (Y) gaps.
    * *Constraint:* You generally move along the axis with the largest gap first.
    * *Format:* "Currently the **[DIRECTION]** direction is the farthest away."
5.  **Stride Selection:** Based on the gap size on that axis, choose FAR, MID, or CLO.

---

**3. RESPONSE FORMAT:**

Reasoning: [Your Step-by-Step Spatial Analysis]
Action: [Single Action Token]

---

**4. EXAMPLES:**

**Example 1: Long Distance Movement**
*Input: Instruction "Open Settings", Cursor at Top-Left, Target (Icon) at Bottom-Right.*
Reasoning: The cursor is currently in the **Top-Left** region. The target 'Settings' is located in the **Bottom-Right** region. More specifically, the cursor is at about [0.1, 0.1] and the target is at about [0.9, 0.9] (relative coordinates) of the image. The target is downwards and to the right of the cursor. Currently the **RIGHT** direction is the farthest away. There is a significant gap. I need a large jump.
Action: <MOVE_RIGHT_FAR>

**Example 2: Micro Adjustment (Same Region)**
*Input: Instruction "Click Search", Cursor slightly above the button.*
Reasoning: The cursor is currently in the **Top-Center** region. The target 'Search' is located in the **Top-Center** region. I need to examine the position more carefully. More specifically, the cursor is at about [0.5, 0.2] and the target is at about [0.5, 0.3] (relative coordinates) of the image. The target is to the **Bottom** of the cursor. Currently the **DOWN** direction is the farthest away. The target is very close. I need a micro-adjustment.
Action: <MOVE_DOWN_CLO>

**Example 3: Execution (On Target)**
*Input: Instruction "Submit Form", Cursor directly on the button.*
Reasoning: The cursor is currently in the **Bottom-Center** region. The target 'Submit' is located in the **Bottom-Center** region. I need to examine the position more carefully. More specifically, the cursor is at about [0.5, 0.8] and the target is at about [0.5, 0.8] (relative coordinates) of the image. The cursor is currently positioned **over** the target 'Submit'. I will perform a click.
Action: <CLICK_SHORT>
\end{lstlisting}

\end{document}